\documentclass[letterpaper]{article} 
\usepackage{aaai24}  
\usepackage{times}  
\usepackage{helvet}  
\usepackage{courier}  
\usepackage[hyphens]{url}  
\usepackage{graphicx} 
\usepackage{natbib}  
\usepackage{caption} 
\usepackage{algorithm}
\usepackage{algorithmic}

\usepackage{newfloat}
\usepackage{listings}
\usepackage{multirow}
\usepackage{booktabs}
\usepackage{amssymb}
\usepackage{amsmath}
\usepackage{subcaption}
\usepackage{bbding}
\usepackage{makecell}
\usepackage[table]{xcolor}
\usepackage{url}
\DeclareCaptionStyle{ruled}{labelfont=normalfont,labelsep=colon,strut=off} 
\floatstyle{ruled}
\newfloat{listing}{tb}{lst}{}
\floatname{listing}{Listing}
\title{SHaRPose: Sparse High-Resolution Representation for Human Pose Estimation}
\author{
    Xiaoqi An\textsuperscript{\rm 1,2},
    Lin Zhao\textsuperscript{\rm 1,2}\footnote{Corresponding authors.},
    Chen Gong\textsuperscript{\rm 1},
    Nannan Wang\textsuperscript{\rm 2},
    Di Wang\textsuperscript{\rm 2},
    Jian Yang\textsuperscript{\rm 1}\footnotemark[1]
}

\affiliations{
    \textsuperscript{\rm 1}PCA Lab, Key Lab of Intelligent Perception and Systems for High-Dimensional Information of Ministry of Education\\
    Jiangsu Key Lab of Image and Video Understanding for Social Security\\
    School of Computer Science and Engineering, Nanjing University of Science and Technology\\
    \textsuperscript{\rm 2} State Key Laboratory of Integrated Services Networks, Xidian University\\
    \{xiaoqi.an, linzhao, chen.gong, csjyang\}@njust.edu.cn, \{nnwang, wangdi\}@xidian.edu.cn
}

\begin{document}

\maketitle

\begin{abstract}
High-resolution representation is essential for achieving good performance in human pose estimation models. To obtain such features, existing works utilize high-resolution input images or fine-grained image tokens. However, this dense high-resolution representation brings a significant computational burden. In this paper, we address the following question: ``Only sparse human keypoint locations are detected for human pose estimation, is it really necessary to describe the whole image in a dense, high-resolution manner?" Based on dynamic transformer models, we propose a framework that only uses Sparse High-resolution Representations for human Pose estimation (SHaRPose). In detail, SHaRPose consists of two stages. At the coarse stage, the relations between image regions and keypoints are dynamically mined while a coarse estimation is generated. Then, a quality predictor is applied to decide whether the coarse estimation results should be refined. At the fine stage, SHaRPose builds sparse high-resolution representations only on the regions related to the keypoints and provides refined high-precision human pose estimations. Extensive experiments demonstrate the outstanding performance of the proposed method. Specifically, compared to the state-of-the-art method ViTPose, our model SHaRPose-Base achieves 77.4 AP (+0.5 AP) on the \textit{COCO} validation set and 76.7 AP (+0.5 AP) on the \textit{COCO} test-dev set, and infers at a speed of $1.4\times$ faster than ViTPose-Base. Code is available at \url{https://github.com/AnxQ/sharpose}.
\end{abstract}

\section{Introduction}
2D human pose estimation (HPE) is a fundamental task in the field of computer vision. Its main goal is to locate a set of anatomical keypoints that correspond to the human body's joints and limbs in an image. HPE has been well studied \cite{guo_pinet,zhang_multi_res_attn,chang_video_hpe_crowed} and forms the foundation for many downstream tasks such as action recognition~\cite{kawai_action_detect,li2017skeleton,xu2022topology,posec3d} and abnormal behavior detection~\cite{tang_abnormal,qiu_abnormal}. Due to its potential applications in the real world, HPE remains an active area of research \cite{niemirepo_bino_rt_3d,yu_lite-hrnet_2021,zhang_fasthumanpose_2019,li_high_perform_hpe,li_onlineknowledgedistillation_2021,jiang_rtmpose}.

In recent years, great progress has been made in human pose estimation~\cite{deeppose,newell_stacked_2016,xiao_simple_2018,hrnet,chen_adversarial_2017}. Most of the leading methods output heatmaps and then take the peak of heatmaps as the keypoint position. Hence, similar to other dense prediction tasks such as semantic segmentation~\cite{transfiner,guo_ISDNet} and depth estimation~\cite{shen_adver_guid,luo_consis_vid_de}, it's necessary to obtain high-resolution representation to ensure the inference accuracy~\cite{badrinarayanan_segnetdeepconvolutional_2017,lin_featurepyramidnetworks_2017,chen_deeplab_2018}. For example, Stacked Hourglass~\cite{newell_stacked_2016} achieves high-quality image representation by stacking a symmetric encoding-decoding structure, while HRNet~\cite{hrnet} utilizes multiple parallel convolution branches to preserve high-resolution feature representations. ViTPose~\cite{xu_vitpose_2022} achieves notable performance using an $8\times8$ fine-grained patch splitting setting.

\begin{figure}[t]
    \centering
    \includegraphics[width=0.9\linewidth]{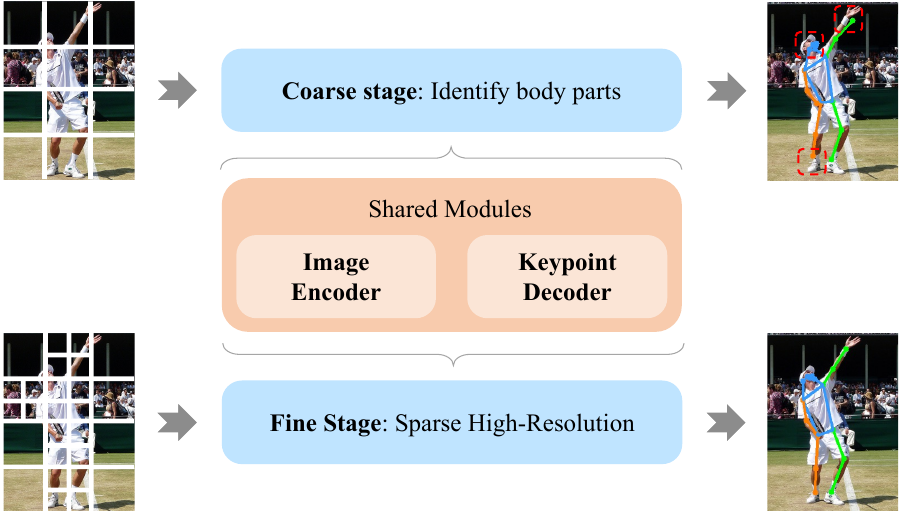}
    \caption{A brief view of SHaRPose. The coarse stage selects image parts contributed to the keypoints, and the fine stage builds high-resolution representations upon them.}
    \label{fig:intro_shr}
\end{figure}

However, it is observed that increasing the resolution of feature representation (\emph{i.e.}, the number of image tokens for transformer-based methods) results in an intensive computational burden. As shown in Table.\ref{tab:comp_cost}, this is particularly significant in Transformer-based methods because the complexity of Transformers is quadratic to the number of tokens~\cite{khan_transformersvisionsurvey_2022}. In this paper, we aim to improve the efficiency of transformer-based models for human pose estimation, and we think about the following question: \textit{Since we only want the keypoint locations, which are sparse relative to the entire image, do we truly need high-resolution feature representation for all contents?}

\begin{table}
  \centering
  \scalebox{0.8}{
    \begin{tabular}{ccccc}
      \toprule
      Model&Input size&FPS&FLOPS&AP\\
      \midrule
      \multirow{2}{*}{HRNet}&256$\times$192 & 194& 15.8& 75.1 \\
                  &384$\times$288 & 152(-21\%)& 35.5(+125\%)& 76.3\\
      \midrule
      \multirow{2}{*}{ViTPose}&256$\times$192& 340& 18.6& 75.8\\
                    &384$\times$288& 143(-58\%)& 44.1(+136\%)& 76.9\\
      \bottomrule
    \end{tabular}
  }
  \caption{Computational cost for high-resolution input}
  \label{tab:comp_cost}
\end{table}

\begin{figure}[tb]
  \centering
  \includegraphics[width=\linewidth]{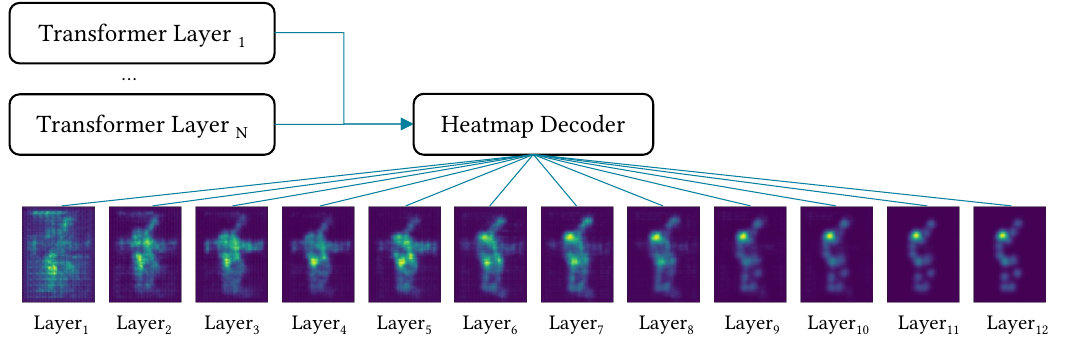}
  \caption{Decoder's response of ViTPose. \textmd{Each heatmap is generated by feeding the output of each intermediate Transformer layer to the heatmap decoder.}}
  \label{fig:act}
\end{figure}

Based on this thinking, we conduct experiments using ViTPose~\cite{xu_vitpose_2022}, as shown in Fig.\ref{fig:act}. Each heatmap is obtained by redirecting the intermediate layer's output to the decoder. These heatmaps provide an intuitive visualization of the image regions that the decoder is focusing on. We can observe that only in the first few layers, the output of the Transformer causes a global response on the decoder, while in the subsequent layers, the decoder's response is clearly concentrated on the sparse local areas containing keypoints. This means that during the inference, a large part of the image tokens like those only containing background information do not provide effective context information. Thus, Focusing solely on keypoint-related image regions may be sufficient to achieve accurate estimation results. And this can significantly reduce the computation costs.

Inspired by this idea, we propose a method that only needs Sparse High-resolution Representation to do human Pose estimation, named SHaRPose. The framework is based on pure transformers and makes use of the correlation mining capabilities of Transformer~\cite{,sebastian_LRP,chefer_trans_inter,liang_evitexpeditingvision_2022} to identify significant image regions for keypoint detection. 

An overview of our framework is illustrated in Fig.\ref{fig:intro_shr}. The inference process is divided into two stages: The initial stage of our network processes coarse-grained patches as inputs, leading to diminished computational expenses owing to the decreased token count. Then a quality predictor module is applied to judge the roughly predicted pose. If the module yields a high confidence score, the network inference terminates. If not, the input image is split into finer-grained patches and fed into the fine stage to get refined results. To avoid computational burden on redundant patches, only the image patches with strong correlations to keypoints are split into finer-grained patches, while patches with weaker correlations are retained in the coarse-grained state. Hence, the proposed approach prevents heavy computational loads caused by processing unnecessary high-resolution image patches.

Overall, the main contributions of this paper are as follows:
\begin{itemize}
  \item SHaRPose proposes to use sparse high-resolution representations, which is the first time that a dynamic optimization strategy has been introduced into the pose estimation task as far as our knowledge goes.
  \item SHaRPose greatly improves the efficiency of pure transformer models in the task of pose estimation. We reduce 25\% of GFLOPs and achieve a $1.4\times$ higher FPS compared to ViTPose-Base.
  \item SHaRPose shows competitive performance with much higher throughput than the existing state-of-the-art models on the MS \textit{COCO} dataset. We achieve 77.4 AP (+0.5 AP) on \textit{COCO} validation set and 76.7 AP (+0.5 AP) on \textit{COCO} test-dev set compared to ViTPose-Base.
\end{itemize}

\section{Related Works}
\subsection{Vision Transformer for Pose Estimation}
Vision Transformers (ViT) crop and map 2D images or image feature representations into token tensors to model long-range dependencies. With the overwhelming performance of Transformers in various computer vision tasks~\cite{dosovitskiy_image_2021,liu_swin_2021,xia_vision_2022,liu_video_2022,carion_end--end_2020,wang_pvt_2021}, some works have introduced Transformers to pose estimation, because the capability of ViT to capture long-range dependencies is of notable value in modeling the structure of the human body~\cite{ramakrishna_posemachinesarticulated_2014,tompson_jointtrainingconvolutional_2014,wei_convolutionalposemachines_2016}. PRTR~\cite{li_prtr_2021} proposes a cascaded transformer structure to achieve end-to-end keypoint coordinate regression. TransPose~\cite{yang_transpose_2021} utilizes a transformer encoder to process feature maps and to produce interpretable heatmap responses. HRFormer~\cite{yuan_hrformer_2021} adopts the structure of HRNet~\cite{hrnet} and inserts attention blocks into branches to achieve larger receptive fields. On the other hand, TFPose~\cite{mao_tfposedirecthuman_2021} uses a set of keypoint queries to regress coordinates from transformers, while TokenPose~\cite{li_tokenpose_2021} proposes token-based heatmap representations to model the body parts explicitly. ViTPose~\cite{xu_vitpose_2022} explores the feasibility of using a plain transformer as the backbone network for pose estimation and achieves excellent prediction accuracy with the help of masked image modeling~\cite{mae} and multi-dataset training.

In general, compared with the pure CNN-based methods~\cite{hrnet,newell_stacked_2016,deeppose}, the transformer-based models are more likely to achieve good results with the help of global attention. However, this also leads to a larger computation cost. In this paper, a sparse high-resolution representation mechanism is explored, which saves considerable computation while retaining the global modeling advantages and high precision of the transformer-based methods.
\begin{figure*}[ht]
    \centering
    \includegraphics[width=0.8\textwidth]{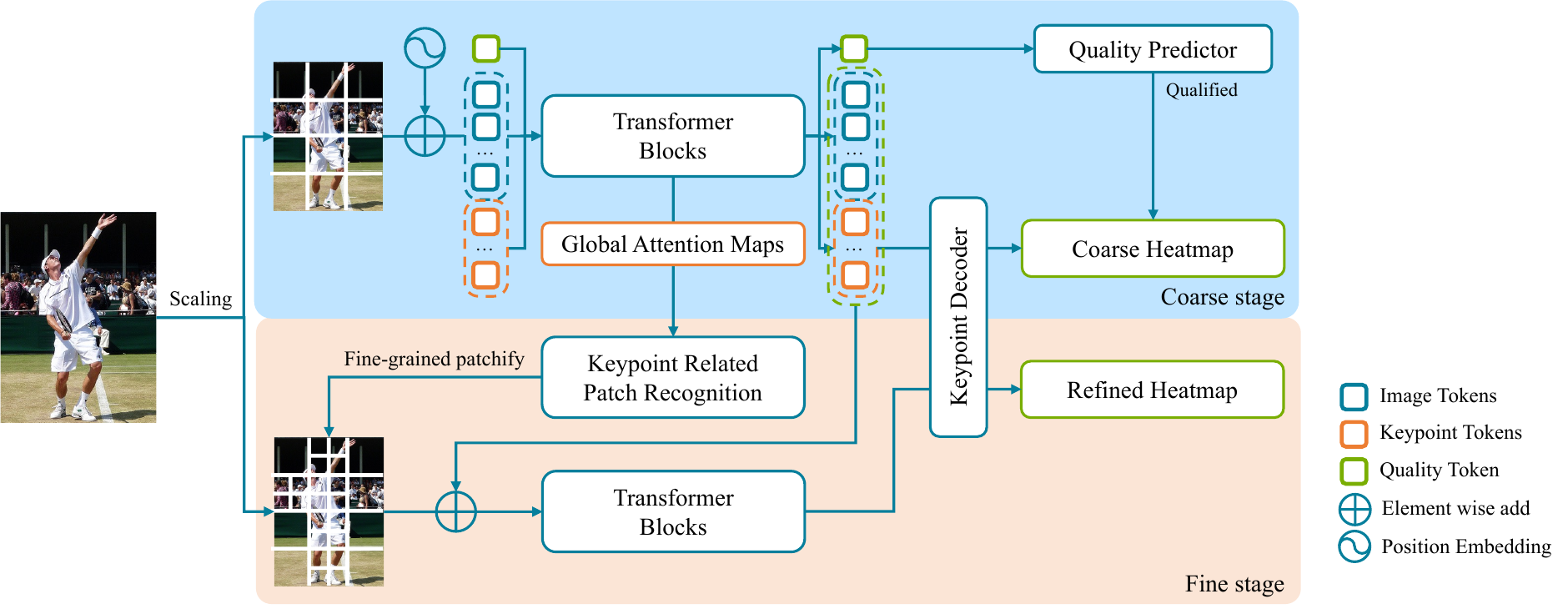}
    \caption{The overall structure of SHaRPose. \textmd{The attention maps yielded by the transformer in the coarse stage is used for selecting keypoint-related patches in the fine stage. Only these keypoint-related patches are processed in finer granularity in the fine stage. The parameters of the Transformer blocks and the keypoint decoder are shared between the two stages.}}
    \label{fig:overall}
\end{figure*}

\subsection{Dynamic Vision Transformer}
To mitigate the issue of computational resource consumption resulting from global feature interaction in Transformers, many methods have been proposed, among which dynamic optimization is one of the major categories.

The simplest approaches involve reducing the number of input tokens to Transformer by pruning them: DynamicViT~\cite{rao_dynamicvit_2021} uses a lightweight detector to determine which tokens to keep, ToMe~\cite{bolya_tokenmergingyour_2023a} fuses similar tokens based on their similarity, and EviT~\cite{liang_evitexpeditingvision_2022} evaluates the importance of image blocks based on class attention. On the other hand, some methods gradually adjust the input granularity from a coarse level. QuadTree~\cite{tang_quadtreeattentionvision_2022a} obtains attention from different scales at each layer and performs cross-scale weighting to capture comprehensive representations, thus reducing the number of tokens involved in attention. DVT~\cite{Wang_2022_CVPR} uses adaptive patch size to reduce the calculation on easy samples. CF-ViT~\cite{chen_cfvitgeneralcoarsetofine_2022} designs two stages using different granularity patches and reorganizes the specific fine-grained tokens with the coarse-grained tokens to refine the prediction in the second stage.

The above-mentioned works have achieved good trade-offs between accuracy and performance. However, the success of these methods is mainly demonstrated in the classification task. In this work, we adapt dynamic transformers to the pose estimation task. Because retaining global context is helpful for human pose estimation, and discarding tokens may cause the model to produce biased predictions, we follow the second category of dynamic transformer methods, designing the framework in a coarse-to-fine manner.

\section{Method}
\subsection{Overall structure}
As depicted in Fig.\ref{fig:overall}, SHaRPose contains two stages with a shared keypoint decoder. The coarse stage consists of a Transformer and a quality predictor module. The fine stage includes a keypoint-related patch selection module and a Transformer sharing the same parameters as the one in the coarse stage.

In this section, we will present our framework stage-by-stage and give a detailed introduction to each module.

\subsection{Coarse-inference stage}\label{sec:coarse_stage}
The goal of this stage is to capture relations between image regions and keypoints, as well as give a coarse inferred heatmap and decide whether the heatmap is accurate enough. To accomplish the objective, a set of keypoint tokens and a quality token are introduced as the queries.

\subsubsection{Token Input}
Denote the input image $X\in\mathbb{R}^{H\times W\times C}$, given the specific patch size $p_h, p_w$ and an input scaling factor $s_c$, we compose the input token sequence as follows:
\begin{equation}\label{eq:coarse_seq}
  \begin{split}
    X_c &= \operatorname{Resample}(X)\in \mathbb{R}^{H\cdot s_c\times W\cdot s_c\times C} \\
    X^c_0 &= \left[v_0^1;v_0^2;\dots v_0^{N_c};k_0^1;k_0^2;\dots k_0^{M};{q_0}\right],
  \end{split}
\end{equation}
where $v_0^i$ is the visual token, obtained from the Re-sampled image $X_c$. First, $X_c$ is split into $N_c=\frac{H\cdot s_c}{p_h}\times\frac{W\cdot s_c}{p_w}$ patches, then a linear projection $f:p\rightarrow v\in\mathbb{R}^D$ is applied to get the corresponding $v_0^i$. $M$ is the number of keypoints, and $\{k_0^i\in\mathbb{R}^D\}_{i=1}^M$ are keypoint tokens from $M$ learnable embeddings, representing the query of keypoints. $q_0\in\mathbb{R}^D$ is a quality token also from a learnable embedding, which will be used to estimate the quality of the predicted human pose.

\subsubsection{Transformer encoder}
After the composition of the input token sequence, a $K$-layers transformer $\mathcal{V}$~\cite{dosovitskiy_image_2021} is applied to obtain the output sequence:
\begin{equation}\label{eq:coarse_tran}
    X^c_K=\mathcal{V}(X^c_0)=\left[v_K^1;v_K^2;\dots v_K^{N_c};k_K^1;k_K^2;\dots k_K^{M};{q_K}\right].
\end{equation}

\begin{figure*}[ht]
  \centering
  \includegraphics[width=0.8\textwidth]{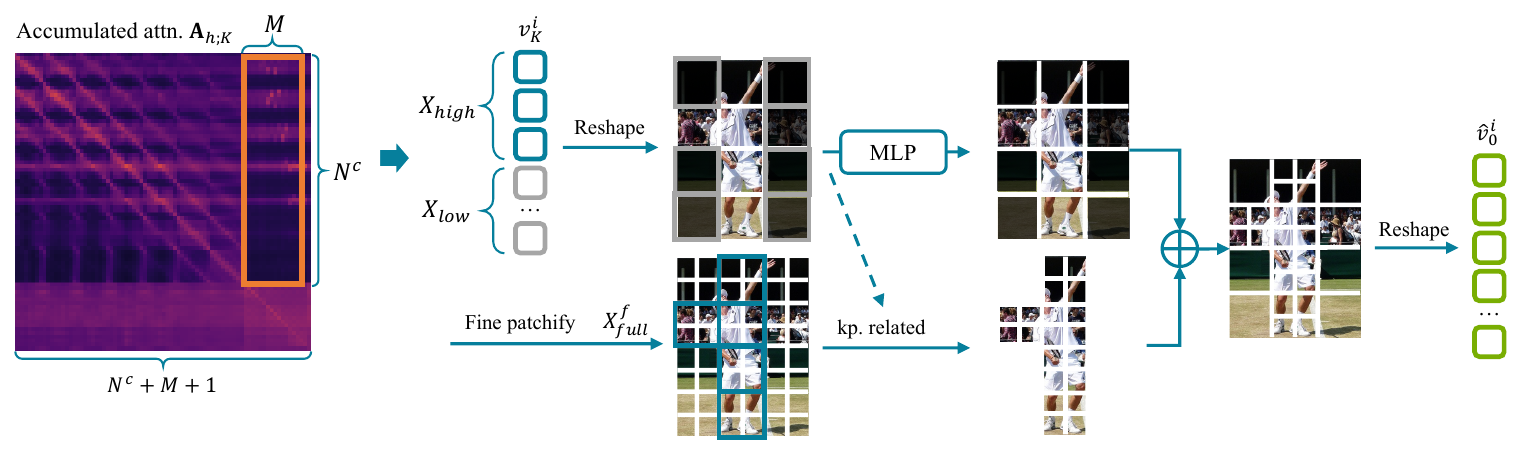}
  \caption{Compose the input of the fine stage. \textmd{The attention scores $\hat{\mathbf{A}}_{h;k}$ between visual tokens and keypoint tokens are just part of the full attention matrix $\mathbf{A}_{h;k}$. Only high-score image patches (blue) are further split into fine-grained patches. An MLP is applied to incorporate the coarse-stage information into the fine stage. }}
  \label{fig:fine_stage}
\end{figure*}

\subsubsection{Keypoint Decoder}
The keypoint decoder builds heatmaps from $M$ output keypoint tokens $\{k_K^i\}_{i=1}^{M}$ through an unified multiple linear projection module:
\begin{equation}\label{eq:keypoint_decoder}
  \mathbf{H}^c_i=\mathcal{D}(k^i_K)\in\mathbb{R}^{\hat{H}\times\hat{W}},
\end{equation}
where $\hat{H}\times\hat{W}$ is the heatmap size, $\hat{H}$ and $\hat{W}$ are both $1/4$ of the original image size $H$ and $W$. Through the decoder, the coarse-predicted heatmaps $\{\mathbf{H}_i^c\}_{i=1}^{M}$ are acquired.

\subsubsection{Quality Predictor}
Inspired by~\cite{zhao_learningacquirequality_2021,pose_nms}, we use a learnable quality embedding $q_0$ to obtain the quality of the predicted keypoints by grubbing information from both visual tokens and keypoint tokens. Then, the quality predictor module produces the quality score of the prediction through the information fused in the quality token:
\begin{equation}
  Q=\operatorname{MLP}(q_K).
\end{equation}
With the estimated quality score $Q$, we set a threshold $Q_{thres}$. Only if $Q<Q_{thres}$, the image will be split into finer-grained patches and processed in the fine stage. This allows the model to dynamically distinguish hard and easy samples. Therefore, the number of images that go through the fine stage can be reduced, which can further increase the throughput.

\subsection{Fine-inference stage}\label{sec:fine_stage}
In this stage, the model generates sparse high-resolution representations and makes high-precision predictions of poses by leveraging the attention obtained in the coarse stage.

\subsubsection{Keypoint-Related Patch Recognition}
In order to decide which image regions need high-resolution feature representations, a kind of relevance score between image patches and keypoints is required. As shown in Fig.\ref{fig:fine_stage}, consider a slice of the attention matrix in a layer of the Transformer $\mathcal{V}$, which is defined as follows:
\begin{equation}
    \hat{\mathbf{A}}_{h;k}=\left[\hat{\mathbf{a}}_{h;k}^{N^c+1} ; \hat{\mathbf{a}}_{h;k}^{N^c+2} ; \ldots ; \hat{\mathbf{a}}_{h;k}^{N^c+M}\right]\in\mathbb{R}^{M\times N^c},
\end{equation}
where $\hat{\mathbf{a}}_{h;k}^{N^c+i}$ is the attention score vector between the keypoint token $k^i_{k-1}$ and all coarse-grained image tokens $\{v^i_{k-1}\}_{i=1}^{N^c}$ at head $h$. This vector reflects the interaction between the keypoint token and the visual tokens. Following \cite{chen_cfvitgeneralcoarsetofine_2022}, we also use exponential moving average (EMA) to combine attentions from each Transformer layer:
\begin{equation}
  \mathbf{\overline{A}}_{h;k}=\beta\cdot\mathbf{\overline{A}}_{h;k-1}+(1-\beta)\cdot\hat{\mathbf{A}}_{h;k},
\end{equation}
in which we set $\beta=0.99$. Then we take the accumulated attention matrix of the last layer $\mathbf{\overline{A}}_{h;K}$ and mix the attention vector of each keypoint and different heads in the following form to get the final visual token correlation score:
\begin{equation}
  \mathbf{s}=\frac{1}{HM}\sum_{h=1}^{H}\sum_{i=1}^{M}\mathbf{\overline{a}}_{h;K}^{N^c+i},
\end{equation}
where $\mathbf{\overline{a}}_{h;K}^{N^c+i}$ is the $i$th column of the matrix $\mathbf{\overline{A}}_{h;K}$, $H$ and $M$ denote the number of heads and the number of keypoints, respectively. According to $\mathbf{s}$, we can rank and select the image patches which are important to estimate human pose. As shown in Fig.\ref{fig:fine_stage}, We select a set $X_{high}$ consisting of $N^h=\lfloor\alpha\cdot N^c\rfloor$ patches with higher scores from $\{v^i_{K}\}_{i=1}^{N^c}$, while the remaining patches form the set $X_{low}$.

\subsubsection{Fine inference}
To perform the fine stage inference, we first need to construct high-resolution representations. Similar to Eq.\ref{eq:coarse_seq} in Section \ref{sec:coarse_stage}, given the scaling ratio in the fine stage $s_f$, the full visual tokens can be obtained as $X_{full}^f=\{v_f^i\}_{i=0}^{N_f}$, where $N_f=\frac{H\cdot s_f}{p_h}\cdot\frac{W\cdot s_f}{p_w}$ is the number of all fine-grained image tokens. 

\begin{table*}[t]
  \center
  \scalebox{0.9}{
    \begin{tabular}{l|cccc|ccc|ccc}
    \toprule
    Model                & Input Size     & Feat. Dim. & Depth & Patch Size   & $s_c$ & $s_f$ & $\alpha$ & FPS & Params  & GFLOPs \\ \midrule
    SHaRPose-Small       & 256$\times$192 & 384            & 12    & 16$\times$16 & 0.5   & 1.0   & 0.5      & 498.3     & 28.4M  & 4.9   \\
    SHaRPose-Small       & 384$\times$288 & 384            & 12    & 16$\times$16 & 0.5   & 1.0   & 0.5      & 395.3     & 48.3M  & 11.0  \\
    SHaRPose-Base        & 256$\times$192 & 768            & 12    & 16$\times$16 & 0.5   & 1.0   & 0.4      & 392.8     & 93.9M  & 17.1  \\
    SHaRPose-Base        & 384$\times$288 & 768            & 12    & 16$\times$16 & 0.5   & 1.0   & 0.3      & 196.6     & 118.1M & 32.9  \\
    \bottomrule 
    \end{tabular}
  }
  \caption{Configurations of the instantiated SHaRPose models. \textmd{We provide the detailed parameters for constructing both the Base and the Small models. And the specific model sizes are presented in the last columns of the table.}}
  \label{tab:implement}
\end{table*}
The initial visual tokens of the fine stage $\{\hat{v}_0^i\}_{i=0}^{\hat{N_f}}$ can be constructed as follows: The first part is composed of tokens not closely associated with keypoints, which can be directly taken from $X_{low}$. The second part comprises tokens generated from $X_{high}$, which are more relevant to keypoints. Denote a singular visual token from $X_{high}$ as $v_K^j$, which is further split into $N=(s_f/s_c)^2$ fine-grained tokens. The computation of the new input visual tokens is formulated as $\{\operatorname{MLP}(v_K^j)+v_f^{c_i}\}_{i=0}^N$, where$\{v_f^{c_i}\}_{i=0}^N$ are fine image tokens from $X_{full}^f$ at the corresponding location of $v_K^j$. Thus, the input token sequence can be formed by:
\begin{equation}
  X^f_0 = \left[\hat{v}_0^1;\hat{v}_0^2;\dots \hat{v}_0^{\hat{N}_f};k_0^1;k_0^2;\dots k_0^{M}\right],
\end{equation}
where $\hat{N}_f=N\cdot\lfloor\alpha\cdot N_c\rfloor+\lfloor(1-\alpha)\cdot N_c\rfloor$ is the number of visual tokens, $k_0^i$ is the same initial keypoint token embedding as in Eq.\ref{eq:coarse_seq}. We present the process of building the fine stage visual tokens in Fig.\ref{fig:fine_stage}.

Then, similar to Eq.\ref{eq:coarse_tran}, a transformer sharing the same parameters as the one in the coarse stage is applied to get the output of the fine stage by $X^f_K=\mathcal{V}(X^f_0)$. Finally, the keypoint tokens are fed into a shared decoder defined in Eq.\ref{eq:keypoint_decoder} to get the fine inferred heatmaps $\mathbf{H}^f_i=\mathcal{D}(k^i_K)$.

\subsection{Loss Function}
For training the network, we impose supervision both on the output heatmaps and the pose confidence that the quality predictor infers:
\begin{equation}
  \mathcal{L}=\mathcal{L}_{heatmap}+\lambda\mathcal{L}_{qp},
\end{equation}
in which $\lambda$ is a hyper-parameter to balance the loss terms. $\mathcal{L}_{heatmap}$ is the heatmap mean square error loss, including the coarse stage and the fine stage:
\begin{equation}
  \mathcal{L}_{heatmap}=\frac{1}{M}\sum_i^M\left(\mathcal{L}_{mse}(\mathbf{H}^c_i,  \mathbf{H}^{gt}_i)+\mathcal{L}_{mse}(\mathbf{H}^f_i,\mathbf{H}^{gt}_i)\right),
\end{equation}
in which $ \mathbf{H}^{gt}_i$ is the ground-truth heatmap. $\mathcal{L}_{qp}$ is an L2-norm loss between the quality predictor's output $Q$ and the coarse stage's ground-truth OKS, which denotes the object keypoint similarity: 
\begin{equation}
  \mathcal{L}_{qp} = \left\Vert Q-\operatorname{OKS}^{gt} \right\Vert_2.
\end{equation}

\section{Experiments}
\subsection{Experiment setup}

\begin{table*}[t]
  \center
  \scalebox{0.9}{
  \setlength{\tabcolsep}{1.5mm}
  \begin{tabular}{l|l|llllll|ll}
  \toprule
  Method                 & Input          & $AP$ & $AP^{50}$ & $AP^{75}$ & $AP^{L}$ & $AP^{M}$ & $AR$ & FPS $\uparrow$ & GFLOPs $\downarrow$ \\ \midrule
  TokenPose-T\cite{li_tokenpose_2021}$\dagger$ & 256$\times$192 & 65.6 & 86.4      & 73.0      & 71.5     & 63.1     & 72.1 & 348.1      & \textbf{1.2}  \\ 
  ViTPose-Small\cite{xu_vitpose_2022}$\dagger$ & 256$\times$192 & \underline{73.8} & 90.3      & 81.3      & 75.8     & 67.1     & 79.1 & \underline{360.3}      & \underline{5.7}  \\ 
  \rowcolor{gray!25}
  SHaRPose-Small$\dagger$& 256$\times$192 & \textbf{74.2}$\uparrow$\small{0.4} & 90.2      & \textbf{81.8}     & \textbf{80.3} & \textbf{71.2}     & \textbf{79.5} & \textbf{498.3}$\uparrow$\small{38\%}      & 4.9$\downarrow$\small{14\%}  \\ \midrule
  SimpleBaseline\cite{xiao_simple_2018}  & 256$\times$192 & 73.6 & 90.4      & 81.8      & 80.1     & 70.1     & 79.1 & 195.1      & 12.8 \\
  HRNet-W48\cite{hrnet}              & 256$\times$192 & 75.1 & 90.6      & 82.2      & 81.8     & 71.5     & 80.4 & 193.5      & 15.8 \\
  HRFormer-Base\cite{yuan_hrformer_2021}          & 256$\times$192 & 75.6 & 90.8      & 82.8      & 82.6     & 71.7     & 80.8 & 122.3      & 14.7 \\
  TokenPose-L/D6\cite{li_tokenpose_2021}         & 256$\times$192 & 75.4 & 90.0      & 81.8      & 82.4     & 71.8     & 80.4 & 348.2      & \textbf{9.9}  \\
  ViTPose-Base\cite{xu_vitpose_2022}& 256$\times$192 & 75.8 & 90.7      & 83.2      & 78.4    & 68.7     & 81.1 & \underline{340.2}      & \underline{18.6} \\
  \rowcolor{gray!25}
  SHaRPose-Base          & 256$\times$192 & 75.5 & 90.6      & 82.3      & 82.2     & \textbf{72.2} & 80.8 & \textbf{392.8}$\uparrow$\small{15\%}      & 17.1$\downarrow$\small{10\%}   \\
  \midrule
  HRNet-W48\cite{hrnet}              & 384$\times$288 & 76.3 & 90.8      & 82.9      & 83.4     & 72.3     & 81.2 & 152.3       & 35.5 \\
  ViTPose-Base\cite{xu_vitpose_2022}& 384$\times$288 & \underline{76.9} & 90.9      & 83.2      & 83.9     & 73.1     & 82.1 & \underline{143.3}    & \underline{44.1} \\ 
  \rowcolor{gray!25}
  SHaRPose-Small$\dagger$& 384$\times$288 & 75.2 & 90.8      & 83.0      & 81.2     & 72.0     & 80.9 & \textbf{395.3}& \textbf{11.0}\\
  \rowcolor{gray!25}
  SHaRPose-Base          & 384$\times$288 & \textbf{77.4}$\uparrow$\small{0.5} & \textbf{91.0}  & \textbf{84.1} & \textbf{84.2}      & \textbf{73.7} & \textbf{82.4}      & 196.6$\uparrow$\small{37\%}        & 32.9$\downarrow$\small{25\%}   \\ \bottomrule
  \end{tabular}
  }
  \caption{Comparison on \textit{COCO} validation set. The same detection results with $56 AP$ are used for human instances. No extra training data is involved for all results. The FPS(frame-per-second) is evaluated under an identical environment. $\dagger$ denotes the small-scale models. The underlined numbers emphasize the compared results. The best results are highlighted in bold. }
  \label{tab:coco_val}
\end{table*}
\begin{table*}[t]
  \center
  \scalebox{0.9}{
  \setlength{\tabcolsep}{1.5mm}
  \begin{tabular}{l|l|llllll|ll}
  \toprule
  Methods               & Input    & $AP$ & $AP^{50}$ & $AP^{75}$ & $AP^{L}$ & $AP^{M}$ & $AR$ & FPS $\uparrow$ & GFLOPs $\downarrow$ \\ \midrule
  SimpleBaseline\cite{xiao_simple_2018}& 384$\times$288 & 73.7 & 91.9      & 81.1      & 70.3     & 80.0     & 79.0 & 153.5      & 28.7   \\
  UDP-HRNet-W48\cite{udp}& 384$\times$288 & 76.5 & 92.7      & 84.0      & 73.0     & 82.4     & 81.6 & 152.3      & 35.5   \\
  DARK-HRNet-W48\cite{zhang_dark_2020}& 384$\times$288 & 76.2 & 92.5      & 83.6      & 72.5     & 82.4     & 81.1 & 150.4      & 32.9   \\
  TokenPose-L/D24\cite{li_tokenpose_2021}& 384$\times$288 & 75.9 & 92.3      & 83.4      & 72.2     & 82.1     & 80.8 & 117.2      & \textbf{22.1}\\
  ViTPose-Base\cite{xu_vitpose_2022}& 384$\times$288 & \underline{76.2} & 92.7      & 83.7      & 72.6     & 82.3     & 81.3 & \underline{143.3}      & \underline{44.1}   \\
  \rowcolor{gray!25}
  SHaRPose-Base         & 384$\times$288 & \textbf{76.7}$\uparrow$\small{0.5} & \textbf{92.8} & \textbf{84.4} & \textbf{73.2} & \textbf{82.6} & \textbf{81.6} & \textbf{196.6}$\uparrow$\small{37\%}  & 32.9$\downarrow$\small{25\%}  \\ 
  \bottomrule
  \end{tabular}
  }
  \caption{Comparison on \textit{COCO} test-dev set, same detection results with $60.9 AP$ is used for human instaces. We only report single dataset training results at resolution $384\times288$.}
  \label{tab:coco_test_dev}
\end{table*}

\subsubsection{Datasets}
We conduct experiments on \textit{COCO}~\cite{coco} and \textit{MPII}~\cite{mpii} datasets. Following the customary strategy of Top-Down methods~\cite{xiao_simple_2018, hrnet, newell_stacked_2016}, we utilize the \textit{COCO} 2017 dataset, which comprises 200k images and 250k person instances. The dataset is segregated into three subsets: train, valid, and test-dev, containing 150k, 5k, and 20k samples, respectively. We train our model on the train subset and test it on the valid and test-dev subsets. The \textit{MPII} dataset, which comprises over 40k person instances and 25k images, is also employed for training and evaluation.

\subsubsection{Evaluation Metrics}
Following~\cite{hrnet, yuan_hrformer_2021, xu_vitpose_2022, li_tokenpose_2021}, we use the standard average precision (AP) as evaluation metric on the \textit{COCO} dataset, which is calculated based on OKS. On the other hand, we perform head-normalized percentage of correct keypoint (PCKh)~\cite{mpii} on the \textit{MPII} dataset and report the PCKh@0.5 score.

\subsubsection{Implementation Details}
The SHaRPose framework  offers variability on three aspects: 1) the embedding size, which specifies the number of features carried by each token; 2) the parameter $\alpha$, which determines the proportion of image patches utilized for generating high-resolution representations; 3) the threshold of the predicted pose quality $Q_{thres}$, which controls the number of samples that enter the fine stage. In this paper, we instantiate SHaRPose with two different sizes by scaling the embedding size. Other configurations like the depth (the number of Transformer blocks) are set the same. The detailed configurations of the instantiated SHaRPose models are presented in Table.\ref{tab:implement}.

\subsubsection{Training Details}
To ensure a fair comparison, all experiments presented in this paper are conducted using the MMPose framework \cite{mmpose} on four NVIDIA RTX 3090 GPUs. The default data pipelines of MMPose are utilized. The masked autoencoder pretrain~\cite{mae} is used as in~\cite{xu_vitpose_2022} for the purpose of exploring the potential capabilities of pure Transformers. UDP~\cite{udp} is used for post-processing. The model is trained for 210 epochs with a learning rate of 5e-4, which is decreased to 5e-5 and 5e-6 at the 170th and 200th epochs, respectively. In particular, we aim to predict the confidence values as accurately as possible with the quality predictor. Because the convergence rate of the quality predictor is much faster than that of the heatmap \cite{zhao_learningacquirequality_2021}, we set $\lambda=0$ in the first 180 epochs and $\lambda=0.03$ in the subsequent epochs based on empirical analysis.

\subsection{Results}\label{sec:result}
\subsubsection{Comparison to state-of-the-art methods on \textit{COCO}}
We compare the performance and efficiency of our proposed method with several state-of-the-art (SOTA) approaches. 

\noindent\textbf{Validation set}\quad
As shown in Table.\ref{tab:coco_val}, under the input resolution $256\times192$, our SHaRPose-Small model achieves an AP of 74.2, which is a significant improvement of +8.6 AP over the TokenPose-T model and a +0.4 AP improvement over ViTPose-Small, while maintaining a faster inference speed. Furthermore, our SHaRPose-Base model achieves an AP of 75.5, which is a +0.4 AP improvement over HRNet-W48 and is $1.9\times$ faster than it. Notably, our model also demonstrates faster inference speed than TokenPose-L/D6, HRFormer, and ViTPose-Base, with comparable accuracy.
At the higher input resolution of $384 \times 288$, our model's advantages become even more pronounced. The SHaRPose-Base model achieves a SOTA performance of 77.4 AP while maintaining lower GLOPs and higher throughput compared to other methods.

\noindent\textbf{Test-dev set}\quad
Table.\ref{tab:coco_test_dev} demonstrates the results of the SOTA methods on \textit{COCO} test-dev. SHaRPose-Base with $384\times288$ as input achieves $76.7AP$. Compared to HRNet with UDP and DARK post-processing, our model achieves +0.2 AP and +0.5 AP higher accuracy and nearly 1.3x faster inference speed. Compared to ViTPose-Base, our model has a +0.5 AP improvement and nearly $1.4\times$ higher throughput.

\subsubsection{Comparison to state-of-the-art methods on \textit{MPII}}
The results on \textit{MPII} test set evaluated by PCKh@0.5 are displayed in Table.\ref{tab:mpii_val}. The input resolution is $256\times256$, and the ground-truth bounding boxes are used by default. Our SHaRPose-Base model achieves a PCKh score of 91.4, outperforming other methods while also demonstrating 2-3 times higher throughput.

\subsection{Ablation Study}
\subsubsection{Influence of $\alpha$}
The parameter $\alpha$ is crucial in controlling the sparsity level of the high-resolution representation, and it impacts the calculation consumed by the fine stage. 

As shown in Table.\ref{tab:alpha}, for 256x192 input resolution, augmenting alpha from 0 to 0.4 can bring significant accuracy improvement, but a marginal gain is observed with subsequent increments. Thus, considering the balance of accuracy and efficiency, we set $\alpha=0.4$. For 384x288 input resolution, increasing $\alpha$ from 0.3 to 0.5 has little effect on accuracy but significantly increases computational costs. Therefore, setting $\alpha$ to 0.3 is sufficient to achieve accurate results.

\begin{table}[t]
  \centering
  \scalebox{0.88}{
  \setlength{\tabcolsep}{1.5mm}
  \begin{tabular}{@{}c|cccc>{\columncolor{gray!25}}c}
    \toprule
    Model & \makecell{Simple\\Baseline} & \makecell{HRNet\\W48} & \makecell{TokenPose\\L/D24} & OKDHP & \makecell{SHaRPose\\Base} \\ \midrule
    Mean$\uparrow$  & 89.0           & 90.1      & 90.2            & 90.6  & \textbf{91.4}          \\
    FPS$\uparrow$   & 66.9           & 47.1      & 65.5            & -     & \textbf{212.4}         \\ \bottomrule
    \end{tabular}
  }
  \caption{Comparison on \textit{MPII} val set. SHaRPose demonstrates a significant advantage.}
  \label{tab:mpii_val}
\end{table}

\begin{table}[t]
  \centering
  \begin{subtable}[t]{0.645\linewidth}
    \scalebox{0.8}{
    \setlength{\tabcolsep}{1.8mm}
      \begin{tabular}{ccc>{\columncolor{gray!25}}ccc@{}}
      \toprule
      $\alpha$ & 0.0  & 0.3  & 0.4  & 0.5  & 1.0  \\ \midrule
      $AP$     & 68.4 & 74.8 & 75.5 & 75.5 & 75.7 \\
      GFLOPs   & 13.3 & 15.8 & 17.1 & 18.2 & 24.9 \\ \bottomrule
      \end{tabular}
    }
    \subcaption{256$\times$192}
  \end{subtable}
  \begin{subtable}[t]{0.345\linewidth}
    \scalebox{0.8}{
    \setlength{\tabcolsep}{1.8mm}
      \begin{tabular}{@{}cc>{\columncolor{gray!25}}c}
      \toprule
      $\alpha$ & 0.3  & 0.5   \\ \midrule
      $AP$     & 77.4 & 77.5  \\
      GFLOPs   & 32.9 & 38.9  \\ \bottomrule
      \end{tabular}
    }
    \subcaption{384$\times$288}
  \end{subtable}
  \caption{The effect of $\alpha$ at different settings }
  \label{tab:alpha}
\end{table}

\begin{figure*}[h]
  \centering
  \includegraphics[width=0.8\textwidth]{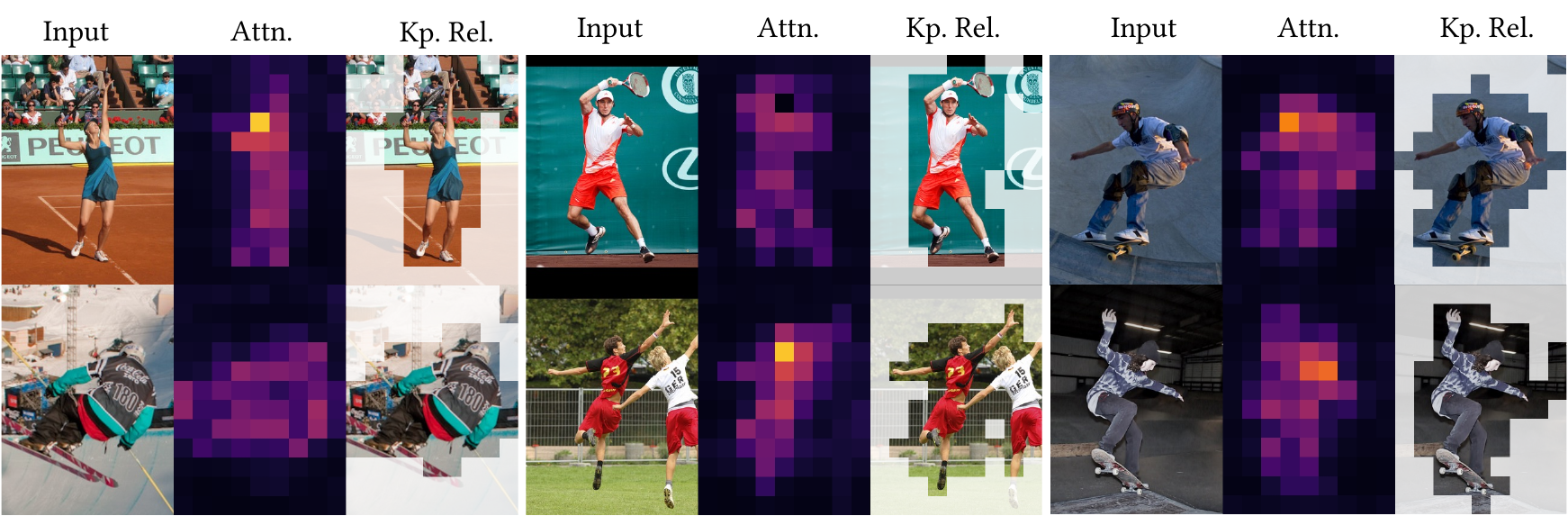}
  \caption{Visualization of keypoint-related regions. Three samples are chosen as examples. The first column gives the input image, the second column presents the accumulated attention map, and the third column shows the selected image regions.}
  \label{fig:vis_attn}
\end{figure*}

\begin{figure}[t]
  \centering
  \includegraphics[width=\linewidth]{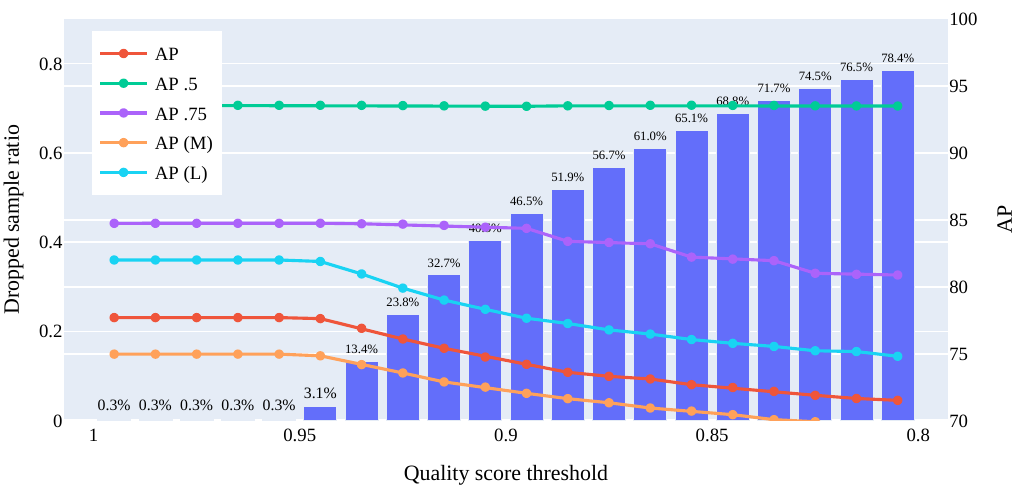}
  \caption{AP and the ratio of dropped samples on different settings of $Q_{thres}$. Each bar demonstrates the ratio of dropped samples with the $Q_{thres}$ given, and the red line denotes the accuracy.}
  \label{fig:drop_frame}
\end{figure}

\begin{table}[t]
  \centering
  \small
  \begin{tabular}{l|cc|cc}
  \toprule
  Method     & $AP$ & $AP^{50}$ & FPS   & GFLOPs \\ \midrule
  DynamicViT & 71.8 & 89.4      & 182.8 & 12.8   \\
  EViT       & 72.7 & 90.1      & 424.1 & 12.8   \\
  \rowcolor{gray!25}
  SHaRPose-S & 74.2 & 90.2      & 498.3 & 4.9    \\
  \rowcolor{gray!25}
  SHaRPose-B & 75.5 & 90.6      & 392.8 & 17.1   \\ \bottomrule
  \end{tabular}
  
  \caption{Comparison with pruning-based dynamic Transformers} 
  \label{tab:comp_prun}
\end{table}

\begin{table}[t]
  \centering
  \scalebox{0.9}{
  \begin{tabular}{l|cc|cc}
  \toprule
  Method            & $AP$ & $AP^{50}$ & FPS   & GFLOPs \\ \midrule
  Coarse-None       & 67.9 & 88.8      & 453.4 & 5.7    \\
  None-Fine         & 74.7 & 90.6      & 302.9 & 17.5   \\
  Coarse-Coarse     & 68.4 & 89.0      & 417.7 & 13.3   \\
  Coarse-NoSel-Fine & 75.7 & 89.0      & 239.5 & 24.9   \\
  \rowcolor{gray!25}
  \textbf{Coarse-Sel-Fine}   & 75.5 & 90.6      & 392.8 & 17.1   \\ \bottomrule
  \end{tabular}
  }
  \caption{Comparison of different configurations of the proposed two stage framework} 
  \label{tab:abl_module}
\end{table}

\subsubsection{Effect of quality predictor}
To evaluate the impact of the quality predictor, we adjust the value of $Q_{thres}$ based on the same SHaRPose-Base model on \textit{COCO} dataset, using the ground-truth bounding boxes. Fig.\ref{fig:drop_frame} illustrates the number of samples that terminate inference after the coarse stage and how the overall AP varies with different values of $Q_{thres}$. We observe that as the value of $Q_{thres}$ decreases, the model tends to skip more samples in the fine stage, resulting in a decrease in $AP$, but the $AP^{50}$ and $AP^{75}$ only change a little. Therefore, the appropriate choice of $Q_{thres}$ depends on the specific application scenario and the required level of accuracy. In the experiments of section~\ref{sec:result}, we set $Q_{thres}=0.95$ for comparison with other SOTA methods.

\subsubsection{Necessity of the coarse-to-fine design}
To demonstrate the necessity of the coarse-to-fine architecture for pose estimation, we analyze from two perspectives: firstly, we perform comparative experiments on two pruning-based dynamic Transformers, namely DynamicViT\cite{rao_dynamicvit_2021} and EViT \cite{liang_evitexpeditingvision_2022}. We introduce keypoint tokens and employ a processing pipeline consistent with SHaRPose. As shown in Table.\ref{tab:comp_prun}, although EViT exhibits higher efficiency, its accuracy is compromised. This indicates that dynamic pruning in localization tasks limits the model's ability to generate precise outcomes. Secondly, we individually remove components of our framework, as shown in Table.\ref{tab:abl_module}. The fine stage is indispensable for accuracy improvement, while the coarse stage, responsible for identifying keypoints-related image patches, plays a crucial role in reducing FLOPs.

\subsection{Visualization}
\subsubsection{Selected keypoint-related image patches}
Fig.\ref{fig:vis_attn} presents some samples to visualize the keypoint-related regions. The second column exhibits the attention map that is accumulated between keypoint tokens and image patches, while the third column shows the keypoint-related regions, which are responsible for generating the high-resolution representation. It can be observed that the attention mechanism is primarily focused on the human instance, which aligns with the original design objective. Moreover, the attention intensity is particularly noticeable on the head since the \textit{COCO} dataset contains more keypoints on the head.

\section{Conclusion}
In this paper, we provide an efficient pose estimation framework using only sparse high-resolution representations, named SHaRPose. Specifically, we introduce token-based keypoint representations into the coarse-to-fine framework to explicitly capture image parts that require high-resolution representations. In addition, we introduce a quality evaluation module, so that the model can quickly complete the inference of simple samples. Our quantitative experiments demonstrate the high accuracy and efficiency of our model. The visualization results also show the effectiveness of the proposed modules. This work provides directions for enhancing the computational efficiency of pose estimation methods using dynamic optimization strategies.

\section{Acknowledgement}
The authors would like to thank the editor and the anonymous reviewers for their critical and constructive comments and suggestions. This work was supported by the National Natural Science Fund of China under Grant No.62172222,62072354, the Postdoctoral Innovative Talent Support Program of China under Grant 2020M681609, and the Fundamental Research Funds for the Central Universities under Grant QTZX23084.
\bibliography{aaai24}
\end{document}